\documentclass[11pt]{article}
\usepackage[margin=1in]{geometry}
\usepackage{amsmath,amssymb,booktabs,graphicx,microtype}
\usepackage{siunitx}
\usepackage{enumitem}
\usepackage[hidelinks]{hyperref}
\usepackage[style=authoryear,backend=biber,maxcitenames=2]{biblatex}
\graphicspath{{figures/}}
\title{\textbf{The Cost of Becoming: Developmental Encoding Increases Phenotypic Diversity but Reduces Locality and Recombination Robustness in Evolved Robots}}
\author{Lyes Saad Saoud\\
Independent Researcher\\
Chicago, Illinois, USA; Abu Dhabi, UAE}
\date{}

\begin{document}
\maketitle

\begin{abstract}
Developmental encodings are often motivated by the expectation that a structured genotype-to-phenotype process can improve evolvability, robustness, and adaptation. Yet an encoding that expands phenotypic variation may simultaneously make useful parental structure harder to preserve under mutation and recombination. We test this trade-off in a controlled evolutionary-robotics benchmark comparing three matched representations: direct encoding, a static generative encoding, and a temporal zygotic developmental encoding. All three conditions use 34 genome parameters, 34 adult-controller parameters, identical initial genome matrices, the same optimizer, the same mutation and crossover operators, the same environments, and the same number of fitness evaluations. Across 30 paired evolutionary runs per representation, developmental encoding does not reliably exceed direct encoding in final training or out-of-distribution fitness. It does, however, substantially increase final population phenotypic diversity and mutant-reachable phenotype diversity. The same representation exhibits lower mutation viability, lower genotype--phenotype locality, sharply reduced crossover offspring fitness and viability, and increased recombination novelty. Continued developmental dynamics also attenuate damage applied early in development relative to otherwise identical late damage. These results identify a representation-level trade-off: development can enlarge the neighborhood of reachable phenotypes and provide within-development damage attenuation while degrading the local inheritance of already-adapted structure. We argue that developmental encodings should therefore be evaluated not by performance alone, but by a joint geometry of generativity, locality, robustness, and inheritance compatibility.
\end{abstract}


\section{Introduction}\begin{figure}[t]
\centering
\includegraphics[width=0.99\linewidth]{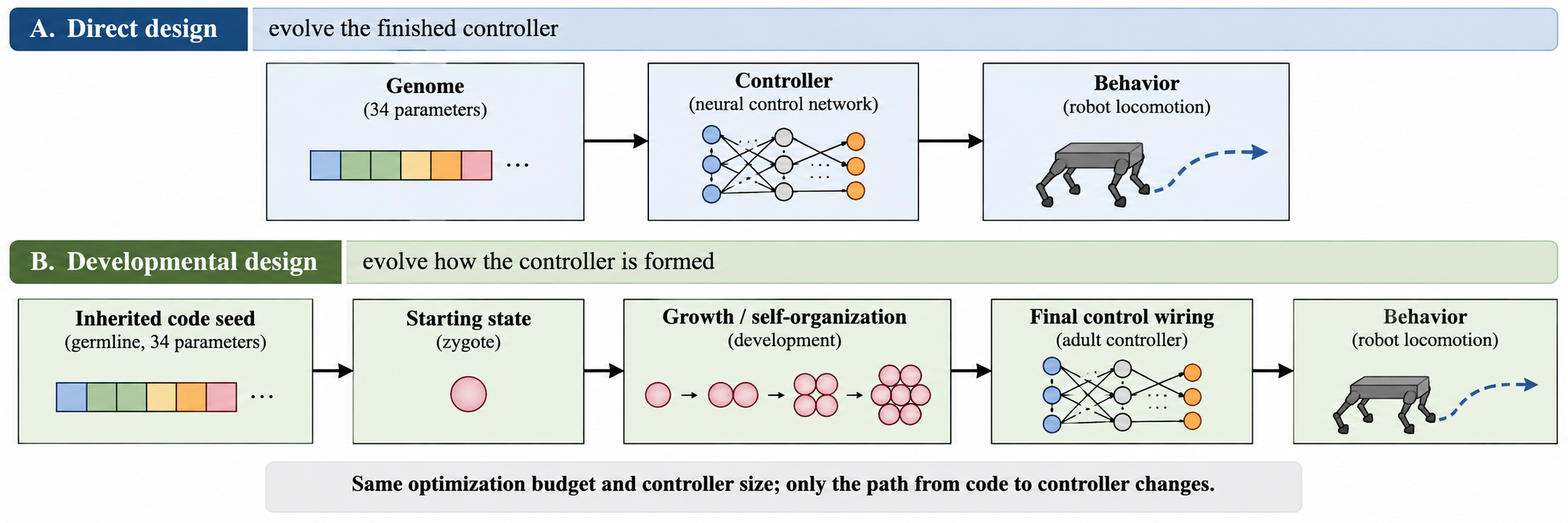}
\caption{Conceptual contrast between direct and zygotic developmental encoding. Direct encoding evolves the finished controller, whereas developmental encoding evolves an inherited code seed that is transformed through a starting state and temporal self-organization before the adult controller is produced. The static generative baseline used in the experiments is omitted here for visual clarity. Genome dimension, mutation, recombination, selection, evaluation budget, and adult-controller dimension are matched across the benchmark; only the genotype-to-phenotype map differs.}
\label{fig:concept}
\end{figure}

Evolutionary robotics depends not only on the optimizer but also on the representation through which inherited variation becomes an embodied controller or morphology. A direct encoding establishes an approximately component-wise map from genome to phenotype. Generative and developmental encodings instead reuse structure, introduce intermediate transformations, or allow temporal dynamics to construct the adult phenotype. Such encodings can express regularity and reuse compactly, but they also reshape the neighborhood structure seen by mutation and recombination \parencite{stanley2003taxonomy,stanley2007cppn,miras2021constrained}.

This distinction matters because ``evolvability'' is not a single property. A representation may generate many phenotypes from nearby genotypes while making most of those variants nonviable. It may increase population diversity while destroying locality. It may tolerate perturbations during development but make sexual recombination more disruptive. Developmental systems can guide evolution under some conditions \parencite{kriegman2018development}, and learned developmental mappings can be explicitly optimized for quality-diversity \parencite{montero2024metalearning}; neither result implies that development universally improves the inheritance of already-adapted structure.

The present study asks a narrower question: \emph{what changes when evolution acts on a temporal developmental map rather than directly on the finished controller, under matched evolutionary budgets?} We focus on controller development in a simulated embodied locomotion system and compare three representation classes under deliberately shared optimizer, dimensionality, initialization, evaluation, and environment conditions.

Our main contribution is empirical rather than terminological. We identify a consistent trade-off between \emph{generativity} and \emph{inheritance compatibility}. Relative to direct encoding, the developmental map yields substantially more phenotypic diversity both at the population level and in local mutant neighborhoods, yet it lowers mutation viability, weakens genotype--phenotype locality, and causes much larger losses under crossover. At the same time, developmental dynamics strongly attenuate perturbations applied sufficiently early that development can continue after damage. The result is not that development is better or worse. It is that development reorganizes where variation appears and how reliably adaptation survives inheritance.

The paper makes four contributions:
\begin{enumerate}[leftmargin=*]
\item a fairness-controlled comparison of direct, static generative, and temporal developmental encodings under equal genome and adult-controller dimensionality;
\item a paired analysis showing that developmental encoding expands phenotype diversity without reliably improving final direct-vs-developmental fitness;
\item evidence that this expanded variation is accompanied by reduced mutation viability, reduced genotype--phenotype locality, and reduced crossover robustness;
\item a timing intervention showing that continued developmental dynamics can attenuate early damage even though the final adult representation is fragile under recombination.
\end{enumerate}

\section{Related Work}
Artificial embryogeny and developmental encoding have long been proposed as ways to reuse genetic information and generate structured phenotypes from compact descriptions \parencite{stanley2003taxonomy}. CPPNs provide a well-known abstraction in which a compact generative mapping produces regular phenotypic patterns without simulating local development \parencite{stanley2007cppn}. This distinction between static generative maps and temporal development motivates the separate baseline used here.

Development can alter evolutionary search by exposing multiple phenotypic states during an individual's ontogeny. In simulated soft robots, morphological development has been shown to guide evolution and promote differential canalization \parencite{kriegman2018development}. Other work has explicitly coupled development and evolution in simulated biorobots or allowed environmental regulation of genotype--phenotype mappings \parencite{long2021embodied,miras2020plasticoding}. More recently, developmental mappings themselves have been meta-learned to increase quality and diversity in generated artifacts \parencite{montero2024metalearning}.

A complementary literature emphasizes that encoding choice imposes strong biases on the phenotype and behavior spaces available to evolution. In particular, low genotype--phenotype locality can make small genetic changes cause large phenotypic changes and can cause high-quality parents to produce poor offspring \parencite{miras2021constrained}. Our study is closest to this representation-centric view. Instead of treating development as an intervention expected to improve fitness, we measure how a temporal developmental map changes performance, local variation, viability, robustness, and recombination at the same time.

\section{Methods}
\subsection{Embodied task and controller}
All individuals control the same simulated articulated locomotor system and are evaluated by the same rollout and fitness functions. The evolved adult controller has 34 parameters in all three conditions. Training uses three fixed environments: \texttt{flat}, \texttt{asym\_left}, and \texttt{noise}. Additional unseen environments are reserved for out-of-distribution evaluation. Fitness combines task displacement, energy use, instability, and recovery with weights frozen before the main evolutionary benchmark:
\begin{equation}
F = 1.00D - 0.28E - 0.83I + 0.79R.
\end{equation}
The same physical state equations, observation process, controller interface, and fitness computation are used for every encoding.

\subsection{Three representation classes}
Each genotype is a 34-dimensional real vector and each adult controller is a 34-dimensional vector. Thus genome size and adult-controller size are matched exactly.

\paragraph{Direct encoding.}
The genome maps directly to bounded controller parameters. There is no intermediate generative computation or temporal developmental process.

\paragraph{Static generative encoding.}
The genome is transformed through a deterministic, non-temporal generative mapping before producing the adult controller. The mapping introduces reuse/nonlinearity but no sequence of developmental states.

\paragraph{Zygotic developmental encoding.}
The genome initializes a temporal developmental system. Repeated local transformations and diffusion-like interactions produce intermediate states before the adult controller is frozen. The genotype-to-phenotype process therefore has a temporal trajectory rather than a single static map.

Figure~\ref{fig:concept} summarizes the distinction.

\subsection{Evolutionary protocol and fairness controls}
The standard benchmark uses population size 64, 90 generations, and 30 independent evolutionary runs per representation. Tournament size is 3, elitism is 5\%, mutation standard deviation is 0.18 with mutation probability 0.35, and recombination probability is 0.6. Every run uses the same optimizer implementation and the same number of fitness evaluations: 5,760 evaluations per representation per run.

Before the benchmark, an automatic fairness audit checked all blocking criteria. The three conditions have identical genome dimension (34), identical nominal genome description length (1,088 bits under the benchmark accounting), identical adult-controller dimension (34), shared mutation, crossover, and selection operators, shared training environments, one shared phenotype-evaluation function, no target or environment argument passed into the developmental function, and paired initial genome matrices across conditions. No blocking check failed.

Two unavoidable asymmetries are declared. First, genotype-to-phenotype compute differs: the direct map uses no developmental multiply-accumulate operations, the static generative map uses 256, and the temporal developmental map uses 3,072 per phenotype construction. Second, equal genome initialization does not imply equal generation-zero phenotype distributions because the maps themselves differ. Both effects are properties of the representation and are reported rather than hidden.

\subsection{Measured outcomes}
We evaluate the representations along six axes.

\paragraph{Evolutionary performance.}
We record best fitness through evolution, final best fitness, evaluations to fixed thresholds, out-of-distribution fitness, and response to an environment switch.

\paragraph{Phenotypic diversity.}
Population phenotype diversity is measured in the common adult phenotype space at the end of evolution. Local reachable diversity is measured by repeatedly mutating an evolved parent and computing dispersion in the resulting phenotype neighborhood.

\paragraph{Mutation viability and locality.}
For each evolved parent, we generate small mutations and record the fraction above a viability threshold. Genotype--phenotype locality is measured by the correlation between genetic distance and phenotype distance in the local mutant cloud.

\paragraph{Recombination compatibility.}
High-performing parents are crossed with the same uniform-crossover operator in every representation. We define
\begin{equation}
R_{\mathrm{cross}} = F_{\mathrm{child}} - \frac{F_A+F_B}{2}.
\end{equation}
We additionally measure offspring viability and phenotype novelty.

\paragraph{Perturbation robustness.}
Adult controllers are evaluated under parameter noise, sensor noise, actuator weakness, and single-module failure.

\paragraph{Developmental damage timing.}
For the developmental representation, three contiguous internal cells are zeroed part-way through development. We compare early damage, when at least eight developmental steps remain, with late damage, when at most two remain. The adult damage baseline applies analogous impairment after the controller is already finalized, when no developmental compensation is possible.

\subsection{Statistical analysis}
The independent evolutionary run is the unit of replication. Because the three representation conditions use matched run indices and paired initialization, the primary direct-versus-developmental contrasts are analyzed as paired data. We report the median paired difference, a 95\% paired bootstrap interval, the Wilcoxon signed-rank test, and matched rank-biserial effect size. Thirty paired runs are available for the main evolutionary, evolvability, robustness, and recombination analyses. The damage-timing comparison also contains 30 paired developmental runs. We do not treat multiple individuals from one evolutionary run as independent replicates.

The original repository also contains a lifetime-learning experiment. A source audit found that its environment-seed derivation used Python's process-randomized \texttt{hash()} function. The source has been corrected, but the expensive experiment has not yet been rerun under that correction; therefore no lifetime-learning result is used in the inferential claims of this paper.

\section{Results}
\subsection{Development does not reliably outperform direct encoding in final fitness}
\begin{figure}[t]
\centering
\includegraphics[width=0.95\linewidth]{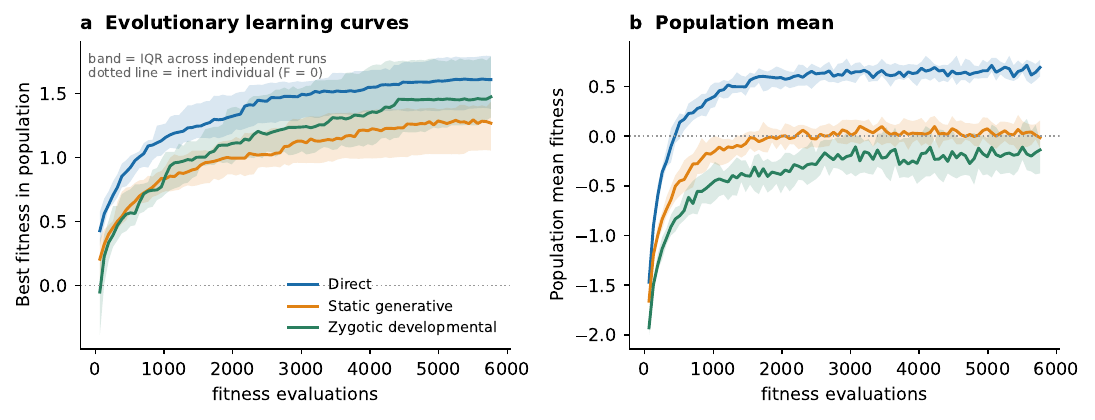}
\caption{Evolutionary learning curves for the three representation classes across matched runs. Development changes the trajectory and variability of search but does not yield a reliable final-fitness advantage over direct encoding in the paired analysis.}
\label{fig:learning}
\end{figure}

Figure~\ref{fig:learning} shows evolutionary trajectories. Median final best fitness was 1.610 for direct encoding, 1.270 for the static generative encoding, and 1.475 for developmental encoding. In the matched direct-versus-developmental comparison, the paired median difference (developmental minus direct) was $-0.231$, with a 95\% paired-bootstrap interval of $[-0.342,\,0.176]$ and Wilcoxon $p=0.213$. Thus the benchmark does not support a claim that temporal development improves final training fitness relative to direct encoding.

The same qualitative restraint applies to overall out-of-distribution performance: the original benchmark did not establish a reliable developmental advantage over direct encoding. Developmental encoding did outperform the static generative baseline on some performance measures, but the central comparison in this paper is direct versus temporal development.

\subsection{Development expands phenotype diversity}
The clearest positive effect of the developmental representation is increased diversity. Final population phenotypic diversity had medians 2.034 for direct encoding and 3.079 for developmental encoding. The paired developmental-minus-direct difference was $+1.067$ with 95\% CI $[0.864,\,1.366]$, Wilcoxon $p=1.86\times10^{-9}$, and matched rank-biserial effect size $1.00$.

The same effect appears locally around evolved parents. Mutant-reachable phenotype diversity had median 1.124 under direct encoding and 1.542 under development. The paired difference was $+0.502$ with 95\% CI $[0.176,\,0.580]$, Wilcoxon $p=5.59\times10^{-9}$, and matched rank-biserial $0.991$.

Thus temporal development substantially enlarges both the diversity retained by evolved populations and the diversity accessible through small genetic perturbations.

\subsection{The diversity gain comes with lower mutation viability and locality}
Expanded mutant diversity did not correspond to more viable variation. Mutation viability fell from a median of 0.906 under direct encoding to 0.648 under development. The paired difference was $-0.273$, 95\% CI $[-0.336,\,-0.172]$, Wilcoxon $p=1.72\times10^{-6}$, with matched rank-biserial effect $-1.00$.

Genotype--phenotype locality also decreased. The median local Pearson correlation between genotype distance and phenotype distance was 0.574 for direct encoding and 0.394 for development. The paired difference was $-0.167$, 95\% CI $[-0.203,\,-0.145]$, Wilcoxon $p=4.66\times10^{-8}$, matched rank-biserial $-0.966$.

\begin{figure}[t]
\centering
\includegraphics[width=0.95\linewidth]{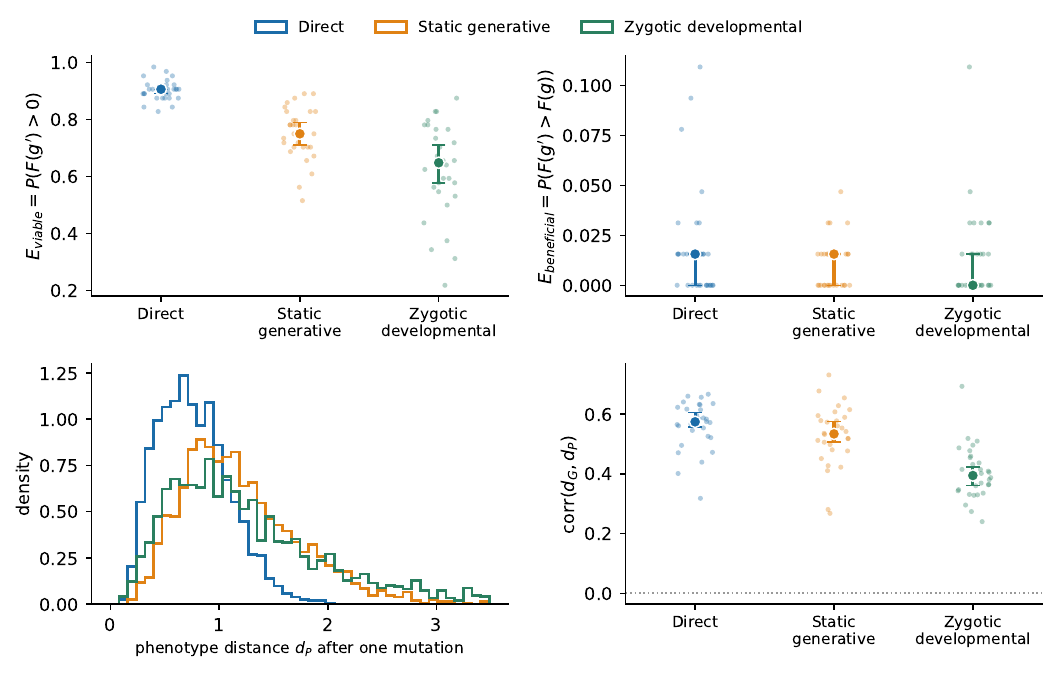}
\caption{Local mutation analysis. Top left: mutation viability, $E_{\mathrm{viable}}=P(F(g')>0)$. Top right: beneficial-mutation probability, $E_{\mathrm{beneficial}}=P(F(g')>F(g))$. Bottom left: distribution of phenotype distance $d_P$ after one mutation. Bottom right: genotype--phenotype locality, $\mathrm{corr}(d_G,d_P)$. In the three summary panels, small points denote individual runs and large markers with error bars denote the median and bootstrap 95\% confidence interval.}
\label{fig:evolvability}
\end{figure}

Figure~\ref{fig:evolvability} summarizes this mixed picture. Development produces a broader local phenotypic neighborhood, but nearby genotypes are less predictably nearby in phenotype space and a smaller fraction remain viable. In this benchmark, ``more variation'' and ``more evolvability'' are therefore not interchangeable statements.

\subsection{Development strongly reduces crossover inheritance quality}
The largest cost appears under recombination. Median $R_{\mathrm{cross}}$ was $-0.161$ for direct encoding and $-0.766$ for developmental encoding. The paired developmental-minus-direct difference was $-0.546$, 95\% CI $[-0.740,\,-0.418]$, Wilcoxon $p=6.15\times10^{-8}$, with matched rank-biserial $-0.961$.

Offspring viability likewise decreased from a median of 0.997 to 0.763. The paired difference was $-0.232$, 95\% CI $[-0.279,\,-0.198]$, Wilcoxon $p=2.84\times10^{-6}$, matched rank-biserial $-0.995$.

At the same time, recombination novelty increased: the developmental offspring were more distant in phenotype space than direct-encoded offspring. The paired novelty increase was $+0.517$, 95\% CI $[0.344,\,0.645]$, Wilcoxon $p=9.31\times10^{-9}$, matched rank-biserial $0.987$.

\begin{figure}[t]
\centering
\includegraphics[width=0.95\linewidth]{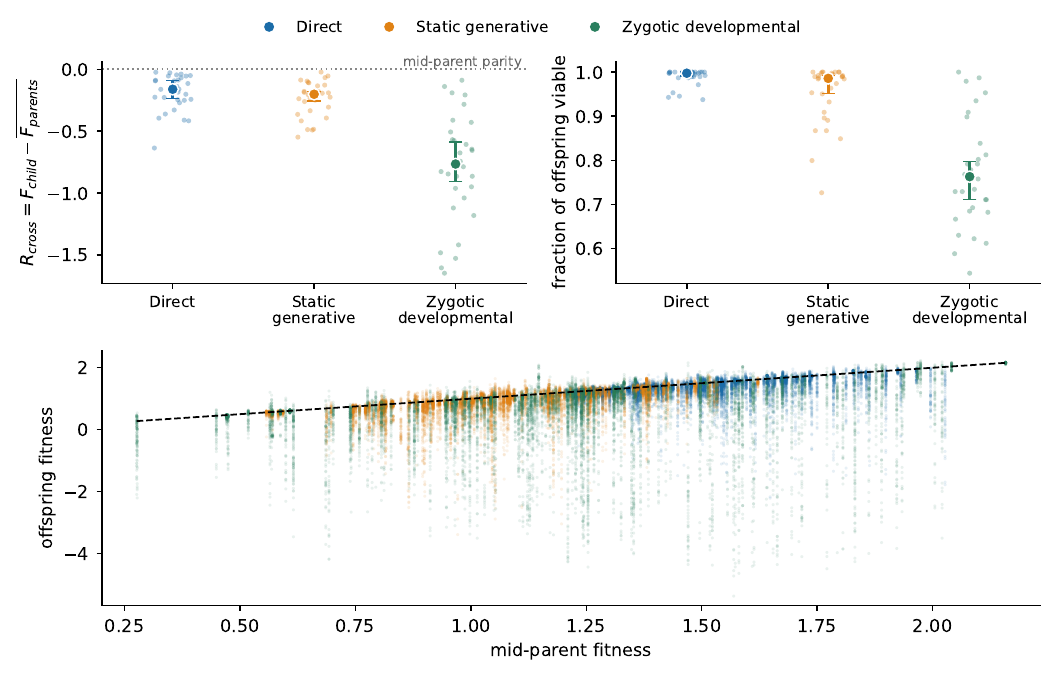}
\caption{Recombination outcomes. Top left: relative crossover fitness, $R_{\mathrm{cross}}=F_{\mathrm{child}}-\overline{F_{\mathrm{parents}}}$; the dotted horizontal line marks mid-parent parity. Top right: fraction of viable offspring. Bottom: offspring fitness versus mid-parent fitness for all recombination trials; the dashed diagonal denotes equal offspring and mid-parent fitness. The developmental representation shows substantially lower crossover fitness and offspring viability.}
\label{fig:recombination}
\end{figure}

Figure~\ref{fig:recombination} therefore exposes the central trade-off in its strongest form. Development makes crossover children more phenotypically novel while making them substantially worse at preserving parental performance. Inheritance becomes more generative and less conservative at the same time.

\subsection{Continued development attenuates early damage}
Development is not uniformly costly. When the same internal damage is applied early enough that developmental dynamics can continue, its adult consequence is dramatically smaller than when applied near the end of development. Median degradation was 0.00125 for early damage and 1.385 for late damage. The paired late-minus-early difference was 1.298, 95\% CI $[1.050,\,1.800]$, Wilcoxon $p=1.86\times10^{-9}$, with matched rank-biserial effect $1.00$.

We interpret this as \emph{damage attenuation by continued developmental dynamics}, not as biological repair. The experiment zeroes internal developmental state and asks whether subsequent dynamics reduce the eventual fitness loss. It does not model tissue regeneration or an explicit repair objective.

Adult module failure also shows a smaller but reliable developmental advantage. Median degradation was 0.491 for direct encoding and 0.363 for developmental encoding; the paired developmental-minus-direct difference was $-0.120$, 95\% CI $[-0.247,\,-0.044]$, Wilcoxon $p=0.00256$.

\begin{figure}[t]
\centering
\includegraphics[width=0.95\linewidth]{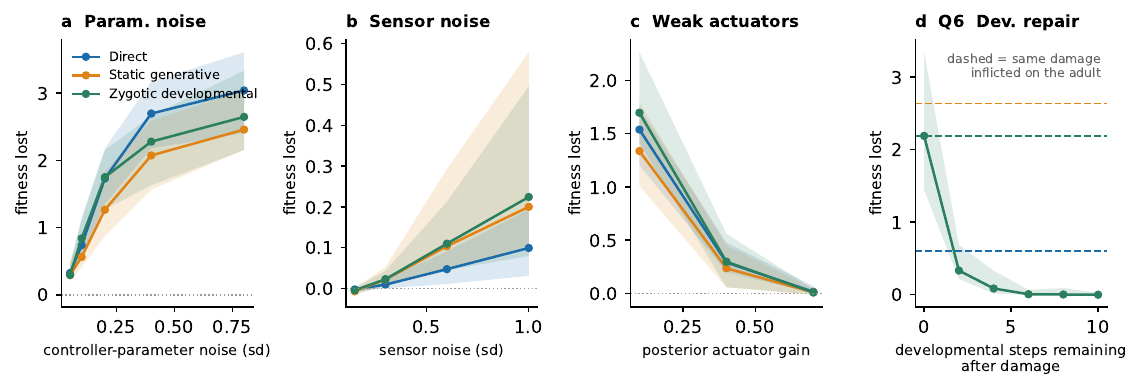}
\caption{Perturbation and developmental-damage analyses. Damage introduced early during the developmental trajectory is strongly attenuated relative to late damage, while adult module failure is also somewhat less costly under the developmental representation.}
\label{fig:damage}
\end{figure}

\subsection{Representation cost}
The comparison intentionally matches genome and adult-controller dimensionality, not genotype-to-phenotype compute. Direct encoding requires no developmental transformation, the static generative mapping uses approximately 256 multiply-accumulate operations per phenotype construction, and the developmental mapping uses approximately 3,072. Across 5,760 evolutionary evaluations per run, this corresponds to 0, 1.47 million, and 17.69 million genotype-to-phenotype operations respectively under the benchmark accounting.

Consequently, the diversity and damage-timing advantages of development are not free. They purchase a different variation geometry at greater mapping cost. Any claim that developmental encoding is ``more efficient'' would therefore require a task in which those structural benefits compensate for the additional compute.

\section{Discussion}
\subsection{The cost of becoming}
The results support a simple interpretation. A temporal developmental map can act as a variation amplifier. Small changes in genotype are transformed through multiple interacting developmental steps, generating adult phenotypes farther apart than those produced by a more direct map. That makes populations and mutant neighborhoods more diverse. But the same amplification lowers locality: nearby genotypes no longer imply nearby adult phenotypes as reliably.

Once sexual recombination is introduced, the consequence becomes especially clear. Crossover combines components from two already-adapted genomes. Under direct encoding, those components map relatively locally to the adult controller. Under the developmental map, recombined components interact through a nonlinear temporal process. The child can therefore become substantially different from either parent's local neighborhood. In our benchmark, that interaction produces higher novelty but sharply lower child fitness and viability.

This suggests separating two notions that are often collapsed under the word ``evolvability.'' We call the first \emph{phenotypic generativity}: the ability of local hereditary variation to access diverse phenotypes. We call the second \emph{inheritance compatibility}: the probability that mutation or recombination preserves enough coordinated structure for the offspring to remain viable and useful. A representation may score highly on one and poorly on the other.

The developmental map studied here occupies exactly that regime: high generativity, low locality, and poor recombination compatibility. This does not make development intrinsically undesirable. It means that developmental encodings may require different variation operators, modularity pressures, mating rules, or evolved mechanisms of developmental buffering than direct encodings.

\subsection{Relation to prior developmental-encoding results}
Our findings are compatible with prior demonstrations that development can guide evolutionary search \parencite{kriegman2018development} and that developmental mappings can be optimized to increase quality-diversity \parencite{montero2024metalearning}. Those works establish that development can provide useful structure under particular regimes. The present study asks a different question: when the representation itself is held to the same genome/controller dimension and subjected to the same mutation and crossover operators, what liabilities accompany increased generativity?

The answer aligns with representation-bias analyses showing that low locality can cause good genotypes to produce poor offspring \parencite{miras2021constrained}. Our contribution is to connect that locality cost directly to temporal development and to separate three simultaneous outcomes: increased reachable diversity, reduced viability/locality, and degraded sexual recombination.

\subsection{Developmental damage attenuation}
The damage-timing experiment provides a counterpoint to the inheritance cost. When perturbation occurs before development is complete, subsequent dynamics can redirect the system toward an adult state with very little loss. The same perturbation applied late causes a much larger deficit. The result demonstrates a temporal property unavailable to a frozen direct map: the trajectory itself can absorb some disturbances.

This should not be confused with an evolved self-repair mechanism. No repair objective was optimized. Rather, the dynamics possess enough convergence or redundancy that an early state perturbation can be partially forgotten before the adult controller is frozen. This property may nevertheless be useful for developmental robotics, where construction noise, component uncertainty, or staged assembly occur before deployment.

\subsection{Implications for Machine Zygote research}
Within the broader Machine Zygote program, this paper shifts the emphasis from whether a developmental germline can generate an offspring to what kind of heredity such a map creates. If development increases novelty while degrading recombination compatibility, then a viable artificial germline may need more than a rich developmental process. It may also need mechanisms that stabilize interfaces between inherited modules, canalize high-value structures, or regulate crossover so that parental modules remain developmentally compatible.

Thus, the next question is not simply how to evolve more complex developmental encodings. It is how to evolve \emph{developmental heredity that is simultaneously generative and composable}.

\section{Limitations}
This study is a simulation and does not demonstrate a physical robot, physical heredity, or biological development. The adult phenotype is a controller parameterization rather than a growing body. The developmental map is one deliberately compact architecture; conclusions should therefore be read as properties of this benchmark class rather than universal laws of development.

The task family is also narrow: one locomotor embodiment, three training environments, and a fixed set of perturbations. Broader claims require replication across morphologies, tasks, and developmental architectures. The three representation classes have matched genome/controller dimension and evaluation count, but not equal genotype-to-phenotype compute; development is substantially more expensive. Initial phenotype distributions also differ because different maps transform the same initial genomes differently.

Finally, the repository's lifetime-learning experiment is excluded from inferential claims pending rerun after correction of a process-dependent seed-generation bug. This exclusion is deliberate: reproducibility of the comparison is more important than retaining an additional result.

\section{Conclusion}
Temporal development changes evolutionary search in a way that is not captured by final fitness alone. In a matched evolutionary-robotics benchmark, zygotic developmental encoding substantially increases population and mutant-reachable phenotype diversity while reducing mutation viability, genotype--phenotype locality, crossover fitness, and offspring viability. Recombined offspring become more novel but less able to preserve parental performance. Continued development can nevertheless attenuate damage that occurs early in the developmental trajectory.

The resulting picture is a trade-off rather than a victory for one encoding class. Development creates a richer space of becoming, but the same nonlinear transformation can make successful structure harder to inherit. For evolutionary robotics, the design target should therefore not be developmental complexity alone. It should be a representation that balances generativity, locality, robustness, and inheritance compatibility.

\section*{Reproducibility and Data Availability}
The complete reproducibility package is publicly available at
\url{https://github.com/LyesSaadSaoud/The-Cost-of-Becoming}.
The repository contains the complete source code, raw and processed CSV files generated by the simulations, evolved-population checkpoints, fairness audit, machine-readable summary, figure-generation scripts, and validation tests. No external empirical dataset is required for the reported benchmark. The standard benchmark uses 30 independent evolutionary runs per encoding, a population size of 64, 90 generations, and a frozen master configuration. The manuscript reports the corrected paired analysis for its primary direct-versus-developmental contrasts. The lifetime-learning experiment is retained in the repository for transparency but is not used as inferential evidence in this manuscript pending a corrected rerun.

\section*{AI-Assistance Disclosure}
AI tools were used to assist with language refinement and editorial polishing. All scientific concepts, experimental design, analyses, interpretation of results, and conclusions were developed and verified by the authors.

\printbibliography
\end{document}